\documentclass{ecai}
\usepackage{graphicx}
\usepackage{latexsym}

\usepackage{algorithm}
\usepackage{algpseudocode}
\usepackage[table,xcdraw]{xcolor}

\usepackage{amsmath}
\usepackage{amssymb}
\usepackage{mathtools}
\usepackage{amsthm}

\usepackage{caption}
\usepackage{multirow}

\ecaisubmission   % inserts page numbers. Use only for submission of paper.
\begin{document}

\begin{frontmatter}

\title{Adversarial Imitation Learning On Aggregated Data}

\author[A,B]{\fnms{Pierre}~\snm{Le Pelletier de Woillemont}}
\author[B]{\fnms{Rémi}~\snm{Labory}}
\author[A]{\fnms{Vincent}~\snm{Corruble}}

\address[A]{Sorbonne Université, CNRS, LIP6, F-75005, Paris, France}
\address[B]{Ubisoft La Forge, France}

\begin{abstract}
Inverse Reinforcement Learning (IRL) learns an optimal policy, given some expert demonstrations, thus avoiding the need for the tedious process of specifying a suitable reward function.
However, current methods are constrained by at least one of the following requirements.
The first one is the need to fully solve a forward Reinforcement Learning (RL) problem in the inner loop of the algorithm, which might be prohibitively expensive in many complex environments.
The second one is the need for full trajectories from the experts, which might not be easily available.
The third one is the assumption that the expert data is homogeneous rather than a collection from various experts or possibly alternative solutions to the same task.
Such constraints make IRL approaches either not scalable or not usable on certain existing systems.
In this work we propose an approach which removes these requirements through a dynamic, adaptive method called \textit{Adversarial Imitation Learning on Aggregated Data} (AILAD). It learns conjointly both a non linear reward function and the associated optimal policy using an adversarial framework. The reward learner only uses aggregated data. Moreover, it generates diverse behaviors producing a distribution over the aggregated data matching that of the experts.
\end{abstract}

\end{frontmatter}

\section{Introduction}

Reinforcement Learning (RL) refers to the process of training an agent to perform a given task through interaction with an environment. It is done by learning a policy which maximizes a given reward signal, or reward function. 
It has demonstrated its capacity to learn optimal, or near-optimal, behaviors in many complex domains \cite{RL_vehicle_navigation} \cite{Atari_DQN} \cite{RL_robotics_survey}, sometimes even outperforming humans \cite{AlphaZero} \cite{AlphaStar}.
However, a reward function which reflects the task well can be hard to develop and maintain.
Moreover, some tasks have conflicting objectives which need to be balanced.
%However it is not without its drawbacks, one of which being the need to design a reward function which reflects the task well. This can be hard to develop and maintain due to the inherent complexity of either the task or the environment itself.
%Some tasks also have conflicting objectives which need to be balanced, or constraints which need to be applied. Correctly specifying the reward function is a challenge.
%For example, in a video game, if the task is to win, then a straightforward reward function yielding a signal at the end of the game, positive if won, negative if lost, would be enough. Now, suppose the task is to win, while also not use half of your ammo, then the reward function would have to reflects that new objective. And so, as environment and task become more complex, so does the reward function. Developing and maintaining a good reward function can be a tedious process.
Imitation learning, in which an agent learns from expert demonstrations,  such as Behavioral Cloning (BC) or Inverse Reinforcement Learning (IRL), are better suited in such situations.
While BC has the appeal of simplicity; it does require large amounts of very fine-grained data to avoid compounding errors, due to the fact that the learned policy influences its future inputs (states) \cite{BC_compounding_error}.
On the other hand, IRL learns a reward function directly, usually from whole trajectories, thus mitigating this compounding error. Most IRL approaches also learn the optimal policy with respect to the learned reward function \cite{MaxEnt_IRL}.
Unfortunately, they often do so by solving the forward problem (finding an optimal policy given the current reward) at each step of the inner loop \cite{IRL_SD}, making such algorithms prohibitively expensive to apply in complex environments.
More recent approaches have managed to alleviate this issue, learning the reward function and the optimal policy conjointly \cite{GCL} \cite{AIRL} \cite{GAIL}. 

However, IRL still suffers from two main limitations: the need for full trajectories and the assumption of homogeneity in the demonstrations.
The need for full trajectories of action-state pairs (or even just states) limits greatly its application, as they might not be readily available.
For instance, if the system relies on third-party hardware which can not be easily manipulated or changed, systems relying on a physical sensors to capture the state, would have incomplete states when the sensors are not working properly. Regulations can also prevent the gathering of such fine-grained data.
% Moreover, most IRL techniques rely on the definition of the states and the actions to be the same for both the reward learner and the RL solver.
%This means that a change in the state definition (e.g. to include missing or new information) would result in the need to gather the expert data once more.
%Moreover, we argue that the requirement to have access to full trajectories is usually a strong constraint on the system for many real-world applications. 
%For example, systems relying on a physical sensors to capture the state, would have incomplete states when the sensors are not working properly.
In addition, storing fine-grained trajectories is more expensive than storing aggregated data (also called \textit{metrics} in the following). If data collection happens remotely, then transmitting full paths might also not be an option due to bandwidth limitations.

Furthermore, most IRL algorithms assume that the expert demonstrations come from a single expert (or from an homogeneous pool of experts), limiting greatly the pool of data from which to train and the capacity of the agent to learn different approaches.
There may be various ways to solve a task, so limiting oneself to a single expert might affect negatively the learning capabilities of the agent. 
For example, each person has its own way to play a video game. Using current IRL techniques to learn how to play a video game would result in the agent learning a single way to play, which might be an \textit{average} behavior corresponding to no existing one.
Indeed, current IRL techniques aim to learn a single behavior, implying an homogeneity in the expert demonstrations.
%IRL techniques aim at finding a unique reward function and its corresponding policy. 
The diversity in the expert trajectories is assumed to be irrelevant for solving the task, like noise. The only way existing techniques take into account this diversity is by learning a probabilistic policy.
% rather than to reproduce the distribution of the experts demonstrations.
%It might also result in the agent learning some \textit{average} behavior which might not be optimal.
%Alternatively, reproducing the diversity of a pool of experts might prove more useful. 
A more realistic assumption may be that trajectories are coming from different experts, and that \textit{the way} they are trying to solve a task is as important as \textit{what the task is}.
Thus, one of the goal should be to reproduce the diversity of the pool of experts.

In this work we try to alleviate these constraints, by training an agent to perform not like one specific expert but like any available expert from a given pool, while only using aggregated data over trajectories.
It builds on previous work \cite{AIRL} \cite{GAIL}, which estimates a non-linear reward function and trains the agent in an adversarial manner.
These approaches have been proven to be equivalent to some degree to the Generative Adversarial Network (GAN) framework \cite{GAN_GCL}. The reward estimator network in IRL can be viewed as the discriminator network in GANs and the agent as the generator.

In our approach, the discriminator network is not trained on state-action pairs, but rather on aggregated data computed at the end of each episode, generated by either the experts or the RL agent.
Moreover, building from both the GAN framework and the goal-conditioned approaches applied to behavioral diversity, we introduce a latent space, given as input to the agent. This latent space, absent from existing IRL approaches, allows the agent to introduce diversity in its behavior, thus producing not a single optimal policy, but rather the full distribution of the experts on the space of aggregated data.
We name this approach Adversarial Imitation Learning on Aggregated Data (AILAD). 
We claim that this is the first working adversarial imitation learning model from aggregated data using a dynamically adaptive non-linear reward function.

%In this paper we first introduce formally the GAN framework and the IRL problem, as well as the link between the two in adversarial settings.
%We also present a video game which serves as our experimental environment and the human data gathered.
AILAD is compared experimentally with CARMI \cite{CARMI}, on the problem of human-like play-style generation.
Play-styles are defined following the work of \cite{PersonasInGame2008}, as a combination of metrics.
The play-style generation problem is a good test bed for our approach since by definition a play-style is measured using aggregated data, which is what our approach relies on.  
Additionally, ablation studies are performed so as to measure the impact of each component.
%We also use this use-case to perform ablation studies so as to measure the effect of each component of our approach.

%Contribution
%use only summary data
%low amount of demonstration
%distribution mapping
%Learns the reward function and the optimal policy jointly, so it does not rely on a MDP for which optima policies can be estimated easily

\section{Background}
\vskip -0.1in
In this section, GAN and IRL methods are formally presented, as well as the link between the two. In addition we present ways to introduce diversity in the behaviors of learned agents.

\subsection{Generative Adversarial Networks}
\vskip -0.1in
Generative adversarial networks, or GANs, are an approach to generate synthetic data matching the distribution of a given dataset, following a certain distribution $p(x)$. It has mostly been applied to computer vision and has yielded impressive results, for example successfully generating photorealistic human faces \cite{GAN_faces}.
In its most basic form \cite{GAN_orignal}, a GAN is composed of a generative model $G$ and a discriminative model $D$. The goal of $D$ is to classify its inputs as either the output of $G$ or as a sample from the underlying distribution $p(x)$. 
The generative model $G$, on the other hand, takes as input some noise $z \sim  \mathcal{Z}$, and aims at generating samples which $D$ would classify as coming from $p(x)$. Its training goal is to maximize the probability that $D$ makes a mistake.
These models usually take the form of neural networks, allowing for the loss of $D$ to backpropagate through $G$.
The original losses for each of the models are the following:
%\begin{equation}
%\mathcal{L}_D(\phi) = \mathbb{E}_{x \sim p}[-\log D_\phi(x)] + \mathbb{E}_{z \sim \mathcal{Z}} [-\log (1-D_\phi(G_\theta(z)))]
%\end{equation}
%\begin{equation}
%\mathcal{L}_G(\theta) = \mathbb{E}_{z \sim \mathcal{Z}}[-\log D_\phi(G_\theta(z))] %+ \mathbb{E}_{z \sim \mathcal{Z}} [\log (1-D_\phi(G_\theta(z)))]
%\end{equation}

\subsection{Maximum Entropy Inverse Reinforcement Learning}
\vskip -0.1in
Our work uses the maximum causal entropy IRL framework \cite{MaxEnt_IRL_PhD}.
Borrowing notations from \cite{AIRL}, an entropy-regularized Markov decision process (MDP) is defined by the tuple $( \mathcal{S}, \mathcal{A}, \mathcal{T}, r, \gamma, \rho_0 )$, with $\mathcal{S}$ the state space, $\mathcal{A}$ the action space, $\mathcal{T}(s'|a,s)$ the transition distribution, $r$ the reward function, $\gamma$ the discount factor and $\rho_0(s)$ the initial state distribution. A sequence of state and action in this environment is called a trajectory, and is denoted $\tau$.

\subsubsection{Forward Reinforcement Learning}
\vskip -0.05in
Finding the policy which maximizes the expected entropy-regularized discounted reward is called \textit{forward} RL. Formally, with $H(\pi(\cdot|s_t) )$ the entropy of policy $\pi$, it tries to solve the optimization problem: $RL(r) = \arg \max _\pi \{\mathbb{E}_{\tau\sim\pi}[\sum_{t=0}^T \gamma^t(r(s_t, a_t) + H(\pi(\cdot|s_t) ) )] \}$.

Trajectories $\tau\sim\pi$ are sampled from the environment using $\pi$. 
The goal is to find the high-entropy policy which maximizes the expected cumulative reward with no prior data, only through trial and error by querying the environment.

\subsubsection{Inverse Reinforcement Learning}
On the other hand, IRL is given prior data $\mathcal{D}_E = \{\tau_1,...,\tau_N  \}$, which is assumed to be coming from a policy $\pi_E$ that is optimal with respect to some unknown reward function $r$. The goal of IRL is to infer $r$, using $\mathcal{D}_E$ and interactions with the environment.
The goal of maximum entropy inverse reinforcement learning is to fit a reward function $r$ from a family of functions $\mathcal{R}$ which satisfies:
$\max_{r \in \mathcal{R}} \{ \mathbb{E}_{\pi_E}[r(\tau)] - \mathbb{E}_{\tau\sim RL(r)}[r(\tau)] \}$.
%$\max_{r \in \mathcal{R}} \{ \mathbb{E}_{\pi_E}[r(\tau)] - [H(RL(r)) + \mathbb{E}_{\tau\sim RL(r)}[r(\tau)]] \}$.
%\begin{eqnarray}
%\max_{r \in \mathcal{R}} \{ \mathbb{E}_{\pi_E}[r(\tau)] - [H(RL(r)) + \mathbb{E}_{\tau\sim RL(r)}[r(\tau)]] \}
%\end{eqnarray}

In other words, the purpose is to find $r \in \mathcal{R}$ that assigns high rewards to the expert policy and low rewards to other policies. A function capable of classifying correctly between expert trajectories and other trajectories would then make a good candidate. This is the purpose of the discriminative $D$ in GAN. The correspondence between IRL and GAN has been explored in details in \cite{GAN_GCL}. Using neural networks as $\mathcal{R}$ and training the RL agent and the reward estimator in an adversarial manner has been the focus of recent work \cite{AIRL} \cite{Options_GAN}. Usually, the discriminative model $D$ corresponds to the reward learner and the generative model $G$ correspond to the learned policy $\pi$.

\subsection{Adversarial Imitation Learning (AIL)}
%\vskip -0.1in
Generative Adversarial Imitation Learning (GAIL) \cite{GAIL} trains $D$ to classify whether a state-action pair $(s,a)$ has been sampled from $\mathcal{D}_E$ or generated using $G$. $G$ on the other hand is an agent,  trained using the TRPO \cite{TRPO} rule, with the reward $r(s,a) = \log D(s,a)$. 
OptionGAN \cite{Options_GAN}, generalizes the approach further assuming that the demonstrations are produced by multiple reward functions, using the option RL framework \cite{Option_RL_framework}. This framework does not interpret the multiple reward functions as concurrent, i.e. various experts solving the same problem, but rather as trying to solve sub-problems.

Given the density $q(x)$ of a fixed generator, the optimal discriminator is \cite{GAN_orignal}:  
$D^*(x) = \frac{p(x)}{p(x)+q(x)}$.
Therefore, one can train $D$ to output this value directly \cite{GAIL}, or if $q$ is easily computable, one can \textit{fill in} the value of $q$ and define $D$ following \cite{GAN_GCL}: 
%%\vskip -0.1in
\begin{equation}\label{eq:GAN_GCL}
D_\phi(x) = \frac{\exp(f_\phi(x))}{\exp(f_\phi(x))+q(x)}
\end{equation}
which would improve the stability of the training. $f_\phi$ is a learned function.
The reward function is defined as $r(x) = \log D(x) - \log(1-D(x))$.  Their proposed approach would have $x \leftarrow \tau$. However, training the reward learner over whole trajectories can result in high variance estimates compared to single state-action tuple ($x \leftarrow (s,a)$). Moreover, computing $\pi(\tau)$ can prove to be expensive. This approach is referred to as GAN-GCL \cite{GAN_GCL}.
%and the \textit{correction} applied in eq.\ref{eq:GAN_GCL} is referred to as the GAN-GCL correction.

Adversarial Inverse Reinforcement Learning (AIRL) \cite{AIRL} applies this GAN-GCL correction to state-action-state $(s,a,s')$ tuple.
Using solely the GAN-GCL correction yields poor results as they show in their work. However, they constrain the shape of $f_\phi$, so as to obtain a reward estimator that is \textit{disentangled} from the environment dynamics:
$f_{\phi, \rho}(s,a,s') = g_\phi(s,a) + \gamma h_\rho(s') - h_\rho(s)$.
This approach, while producing good theoretical and empirical results, relies like most IRL techniques on the availability of whole trajectories.
It makes these approaches poorly suited to many real-world problems where only aggregated data is available, due to third-party constraints or to sensors malfunction or even bandwidth availability.

Moreover, these adversarial imitation learning algorithms (AIL) still suffer from the homogeneity assumption discussed above.
The assumption of homogeneity in expert behavior in IRL is reflected in the formulation of these approaches.
In MaxEnt IRL \cite{MaxEnt_IRL} for instance, the experts can generate suboptimal trajectories, but the probability of doing so exponentially declines as the cost of the trajectories escalates. These approaches regard sub-optimal demonstrations as trajectories which failed to maximize a unique underlying reward function.
Our approach regards these demonstrations as trajectories which successfully maximized their specific reward function. 

%Another limitation of these adversarial imitation learning algorithms is the assumption that the expert demonstrations are coming from a single expert (or an homogeneous pool of experts).
%A more realistic assumption should be that trajectories are coming from different experts, all solving the same tasks in different ways -- or possibly from a single expert, behaving in various ways on the same task. Not accounting for that diversity in the expert data makes these approaches unpractical in real-world settings.

Furthermore, our approach relies solely on aggregated data, and takes into account the potential diversity in the behaviors that produced it. Indeed we introduce a noise vector as additional input to our agent, similar to the latent space vector in GANs, or the goal embeddings in goal-conditioned RL. This allows the agent to produce more diversity.
Both GAIL and AIRL work on state-action pairs, not trajectories, while our approach uses aggregated data over whole trajectories. Our approach is close to GAN-GCL, it is important to note that GAN-GCL, on state action pairs performs quite poorly, as demonstrated in \cite{AIRL} and that no experiments have been performed using GAN-GCL on trajectories.
Moreover, simply using GAN-GCL architecture with aggregated data would still perform worse than our approach, as is demonstrated in the later ablation study.

\subsection{IRL On Aggregated Data}
%\vskip -0.075in
The requirement for full paths can be a limiting factor, as they might not be available. Some approach have been put forth to try and overcome this. 
For instance, \cite{IRL_original} assume that instead of full trajectories, the expected sum of discounted state feature values that the agent encounters during the demonstrated paths is observed.
%Observing correctly the state at each time-step might not be a reasonable expectation. Instead, \cite{IRL_stochastic_obs} assume the observation of the state to be probabilistic instead. relaxing the need for state observations to always be accurate.
%These approaches however still impose strong constraints on the system which collects the data.
Expanding on this notion of features, many MaxEnt IRL techniques rely on feature counts \cite{MaxEnt_IRL}. 
However, feature counts are sums of features (typically simple binary functions), over trajectories  and the relationship between these feature counts is typically linear. Technique using neural network \cite{wulfmeier2015maximum} rely on simple domains where analytically solving the problem with value iteration is tractable.
Other techniques relying only on summary data, aggregated over the trajectory rely on having an MDP simple enough to be solved in the inner loop of the algorithm \cite{IRL_SD}. This requirement for a RL solver in the inner loop makes this approach impractical in many complex environments where solving any MDP can prove too costly to be done at high frequencies.

\subsection{Diversity And Distribution Mapping}
%\vskip -0.1in
As discussed above, the assumption of homogeneity in the experts demonstrations limits greatly the applications of existing IRL methods.
%The other main limitation of existing IRL methods is the assumption of homogeneity in the demonstrations. Indeed, current methods aim to find the reward function which led to these demonstrations.
%This implies that all demonstrations maximize the same reward function. In most real-world applications, the demonstrations are not coming from a single agent, but rather from humans and often from more than one person: it is fair to assume that these demonstrations are heterogeneous.
%Moreover, even if the demonstrations came from a single person, they might solve the same problem in different ways. 
Therefore, diversity ought to be infused in the agent.
Diversity for the sake of diversity can be achieved in different ways, one is through the Quality-Diversity framework \cite{Map_Elite}. 
However, the goal of our work is to create an agent capable of reproducing the distribution of behaviors produced by the experts, as measured by the aggregated data summarizing their behaviors.

%So the desired diversity should be contained in this distribution.
%However, in our case we are not necessarily looking for diversity but rather for heterogeneity. In this work we are trying to map a given distribution, that is to create a very specific and aimed diversity.

Two approaches can be used to that effect: goal-conditioned agents or stochastic generative agents.
A recent approach called G-Flow Network \cite{FlowNetwork} is a good example of the latter. This approach trains an agent that will produce a trajectory with a probability that is proportional to the reward associated with it.
Goal-conditioned agents can also be used to create diversity. A goal conditioned agent \cite{goal_cond_rl_old} is an agent characterised by a policy $\pi(a|s,z)$, where $z$ is some additional information given to the agent.
One recent example of this is the Configurable Agent with Reward as Input (CARI) \cite{CARI_paper} approach. In it,  $z$ represents the weights of the linear reward function, thus creating a single agent capable of generating a continuum of diverse behaviors, by simply sampling various values of $z$ during inference.

Another approach called Configurable Agent with Relative Metrics as Input (CARMI) \cite{CARMI} uses $z$ to represent the metrics that the agent should produce in order to gather the maximum reward.
If the metrics come from humans and $z$ is drawn randomly from the distribution of those metrics, then the agent is capable to generate any human-like metrics. They apply their approach to automated video game testing.
This approach is the closest to the one presented in this paper in terms of both application and objective, and thus will serve as the baseline in our experiments.
Their approach relies on  a linear reward function.
The agent receives a negative reward, proportional to the distance between its goal $z$ and the metrics it generated. Both its goal and the metrics generated are each normalized using the experts distribution. Such a distance treats all the metrics the same way.
A linear, un-weighted, relationship between metrics is a strong assumption to make.
For example, consider a self-driving car with 10 available metrics, 9 of which are speed related and only 1 being safety related. An agent could learn a policy providing good returns by focusing solely on speed. 
Moreover, the reward function in CARMI does not take into account the correlation between all metrics. This can create sub-optimal behaviors, in which the agent only focuses on some metrics and gives average results on others.
Finally, for CARMI to train well and be useful at inference, one needs an effective sampling strategy for the targeted metrics, this usually requires either expert knowledge (e.g. how to weight the metrics) or extensive preprocessing (e.g. clustering), which can sometimes introduce undesired bias in the resulting behaviors. Our approach suffers from none of these limitations.

Our approach could be viewed as a goal-conditioned agent: the agent's policy is indeed of the form $\pi(a|s,z)$. However, in contrast with CARMI, here $z$ does not have any obvious interpretation, as is usually the case with latent spaces, in the same way that the noise given as input to the generator in a GAN has no direct interpretation. Indeed, here $z$ is only there to allow the agent to create diverse solutions, without having to maintain a sub-optimal high entropy.

Table \ref{tab:approach_comparatif} compares the requirements of the main approaches presented above. This shows that there are approaches to overcome one or more of the four limitations presented in this table. However, none exists to overcome them all. 
%\vskip -0.2in
\begin{table}
\caption{Comparison of the requirements for the main imitation learning approaches.}{}
\label{tab:approach_comparatif}
%\vskip 0.15in
\begin{center}
\setlength\tabcolsep{2.5pt}
\begin{small}
%\begin{sc}
\begin{tabular}{c|cccc}
       & \begin{tabular}[c]{@{}c@{}}Full \\ trajectories\end{tabular} & \begin{tabular}[c]{@{}c@{}}Experts\\ homogeneity\end{tabular} & \begin{tabular}[c]{@{}c@{}}Linear\\ reward\\ function\end{tabular} & \begin{tabular}[c]{@{}c@{}}Solving forward\\ RL at high\\ frequency\end{tabular} \\ \hline
AIL    & {\color[HTML]{FF0000} Yes}                                   & {\color[HTML]{FF0000} Yes}                                    & {\color[HTML]{70AD47} No}                                          & {\color[HTML]{70AD47} No}                                                        \\ \hline
IRL-SD & {\color[HTML]{70AD47} No}                                    & {\color[HTML]{FF0000} Yes}                                    & {\color[HTML]{70AD47} No}                                          & {\color[HTML]{FF0000} Yes}                                                       \\ \hline
CARMI  & {\color[HTML]{70AD47} No}                                    & {\color[HTML]{70AD47} No}                                     & {\color[HTML]{FF0000} Yes}                                         & {\color[HTML]{70AD47} No}                                                        \\ \hline
AILAD  & {\color[HTML]{70AD47} No}                                    & {\color[HTML]{70AD47} No}                                     & {\color[HTML]{70AD47} No}                                          & {\color[HTML]{70AD47} No}                                                       
\end{tabular}
%\end{sc}
\end{small}
\end{center}
%\vspace{-9mm}
\vskip -0.55in
\end{table}

\section{Adversarial Imitation Learning on Aggregated Data}
%\vskip -0.15in
In this section, the aggregated data, the model and the training procedure are presented in details, as well as the inference strategy used in the experiments.
%In this section we introduce formally the summary data as well as our model and the training procedure, as well as the inference strategy used in our experiments.
This approach learns conjointly the non-linear reward function and the optimal agent in an adversarial setup, solely based on data aggregated over trajectories (i.e. metrics), rather than full trajectories. Moreover, we introduce a noisy latent space, allowing the agent to create diversity, thus removing the underlying need for homogeneity in the expert data. We call this approach Adversarial Imitation Learning on Aggregated Data (AILAD).

\subsection{Proposition}
%\vskip -0.05in
\subsubsection{Aggregated Data}
As stated before, we assume that full expert trajectories are not available, instead the agent can access data aggregated over each trajectory. Formally, we do not have access to $\mathcal{D}_E = \{ \tau_1^{(E)}, ..., \tau_N^{(E)} \}$,
but rather to 
$\mathcal{D}_{\psi,E} = \{ \psi(\tau_1^{(E)}), ..., \psi(\tau_N^{(E)}) \}$
where $\psi$ is called the summarizing function \cite{IRL_SD}: a function which takes as input a trajectory and outputs $M$ aggregated metrics, $\psi(\tau) \in \mathbb{R}^M$.
For a navigation agent, the metrics could for example be: the average speed, the average acceleration, the number of brakes hits.
To simplify matters, the function $\psi$ is also assumed to normalize the data with respect to the distribution of $\mathcal{D}_{\psi,E}$, assuming a Gaussian distribution. We will use $\psi_\mathcal{N}$ when we need to explicitly mention it, otherwise $\psi$ will be used to simplify notations.

The summarizing function is interpreted in our work as a " low-dimension mapping from state-actions pairs to lower-dimension representation", thus resulting in a set of metrics. In certain real-world systems, only the set of metrics is observable in the expert dataset, not the state-actions pairs. The transformation from trajectories to metrics is only used for the agent during training, but not for the experts' data. In other words, we do not gather expert trajectories which are then transformed into metrics, rather we gather expert metrics directly. Meaning that our approach relies solely on metrics from the expert, not detailed trajectories. 

\subsubsection{Reward Learner And Reward Function}
Our approach is related to GAN-GCL \cite{GAN_GCL}, see eq.\ref{eq:GAN_GCL}, but instead of $x \leftarrow \tau$, we use $x \leftarrow \psi(\tau)$. 
Because our approach relies on aggregated data and not full paths, computing $q$ is quite straightforward, and requires little computational power.
The reward learner $D_\phi$ is a neural network which takes as input a series of metrics and outputs the probability that those metrics came from $\mathcal{D}_{\psi,E}$.
The reward learner output takes the form
\begin{eqnarray}\label{eq:GAN_R_D_GCL}
D_\phi(\psi(\tau)) = \frac{\exp(f_\phi(\psi(\tau)))}{\exp(f_\phi(\psi(\tau))+\pi_\theta(\psi(\tau))}
\end{eqnarray}
where $\pi_\theta(\psi(\tau))$ is the RL agent's distribution over the metrics space -- it is pre-computed and its value \textit{filled in}. 
$\pi_\theta(\psi(\tau))$ is computed using the data generated by the agent on previous trajectories. This represents the distribution of the agent, in the space of the game metrics. During training, in addition to the usual RL replay buffer, we store the metrics generated by the agent to estimate this value. This represents the probability that the agent $\phi_\theta$ generates the metrics $\psi(\tau)$.
The discriminative model's parameters are only those of $f_\phi$. Using this GAN-GCL correction creates a fixed convergence point for $D_\phi$, which translates into a more stable reward function for the generator.

Using a neural network allows the learner to represent the reward as any non-linear combination of the metrics.
Thus, there is no need to specify how much each component should matter with respect to the others.
%, or even how they should be weighted. 
It is trained using positive example coming from expert data $\mathcal{D}_{\psi,E}$ and negative example sampled from the environment, using $\pi_\theta$.

The generator, or $\pi_\theta$, is trained using any RL solver algorithm using the reward function
\begin{eqnarray}\label{eq:reward_formula}
r(\tau) = \log D_\phi(\psi(\tau)) - \log(1-D_\phi(\psi(\tau)))
\end{eqnarray}
This and the non-linearity of $D_\phi$ means that the agent has to produce metrics similar to $\mathcal{D}_{\psi,E}$, on each dimension. If the agent fails to match $\mathcal{D}_{\psi,E}$ on even one dimension, then the discriminator can just focus on that single dimension and the agent will only get low rewards, even though it is behaving well on every other dimensions of the aggregated data. 
Moreover, the agent must produce metrics that respect the underlying correlations present in the expert dataset.

\subsubsection{Latent Space}
In addition to training the reward model on aggregated data only, we introduce a latent space $z \sim \mathcal{Z}$. This latent space serves the same purpose as in the GAN framework: generate diversity. At the beginning of each episode a new $z \sim \mathcal{Z}$ is randomly sampled, and given as input to the agent $\pi(a|s,z) =: \pi_z(a|s)$. 
%Doing so helps the RL agent to introduce diversity in its behavior.
Without it, the only diversity observed would be the one brought in by the stochasticity of the environment or the policy, limiting greatly the expressiveness of the available diversity in the demonstrations. For simplicity $\mathcal{Z} = \mathcal{N}(0, \mathbb{I}^M)$. In goal-conditioned RL, $z$ is also reflected in the reward function, here the discriminator. In this work, it is not the case and should be the focus of future work.
The models could very well ignore $z$. However, the ablation study performed, wherein $z$ is removed, shows otherwise, with the resulting distribution lacking diversity. 
%This should be an area of future work, so as to improve the diversity of generated behaviors of the agent.
The overall training procedure is presented in Algorithm \ref{alg:AILAD}.

%\vskip -0.2in
\begin{algorithm}[h]
\caption{Adversarial Imitation Learning on Aggregated Data}\label{alg:AILAD}
\begin{algorithmic}[1]
\Require $D_{Epcohs} \in [0;1]$ and $K$
\State Obtain expert data $\mathcal{D}_{\psi, E}$
\State Initialize $\pi_\theta$ and $D_\phi$. Set $\mathcal{D}_{\pi} = \{\}$
\State Pre-train $D_\phi$ using $\mathcal{D}_{\psi, E}$ and synthetic data (Sec. \ref{sec:D_pretrain})
\For{$k$ in $[1,K]$}
    \State Draw $z \sim \mathcal{Z}$ and collect $\tau_i$ using $\pi_{\theta,z}$
    \State Put $\tau_i$ in $\mathcal{D}_{\pi}$
    \State Compute $\pi_\theta(\psi(\cdot))$ using $\mathcal{D}_{\pi}$ \Comment{For eq. \ref{eq:GAN_R_D_GCL}}
    \If{random() $< D_{Epcohs}$}
        \State Update $D_\phi$ to differentiate $\mathcal{D}_{\pi}$ from $\mathcal{D}_{\psi, E}$
    \EndIf
    \State Update $\pi_\theta$ using any $\tau \in \mathcal{D}_{\pi}$ and $r(\tau)$ (eq. \ref{eq:reward_formula})
\EndFor
\end{algorithmic}
\end{algorithm}

\subsection{Stability Improvements}\label{sec:stbaility_inprov}
\vskip -0.05in
Because RL algorithms and GAN trainings usually suffer from high variance, it is important to specify the stability improvements put in place.
Each component developed below is evaluated in the ablation study in section \ref{sec:ablation_studies_results}.

\subsubsection{Pre-Training of D}\label{sec:D_pretrain}
The beginning of the training is crucial in many RL approaches. Indeed, the first trajectories could very well guide the agent into a local minima, by creating a vicious circle.
In AILAD, the first trajectories also impact the reward function estimation.
No pre-training $D$ leads to having a random reward function at the beginning of the training.
Optimizing on a random reward function can produce a vicious loop, where the agent performs nothing like the experts, providing no example of high rewards as the discriminator can easily distinguish between the agent's metrics and the experts'. And so, without the proper exploration strategy, the agent will most likely never create any expert-like data.
To avoid this, the discriminator model is pre-trained, so that it can give meaningful feedback to the generator right at the beginning of training.

%It is fair to assume that each normalized metric rests in a $[-3; +3]$ interval. Thus synthetic data can easily be created by sampling randomly in $[-3; +3]^M$, and using those as fake negative examples.
%Another way to generate this synthetic data would be to run a random action heuristic in the environment. We opted for the synthetic data approach in order to reduce the cost of interacting with the environment, which in some cases might be too costly.

As for the \textit{filled in} value of $\pi$ in eq. \ref{eq:GAN_R_D_GCL}, a simple constant is used. Using a high value is preferred. 
%Doing so forces $D_\phi$ to fit $f_\phi$
By doing so, when presented with the actual $\pi$ distribution during the training, $D$ will start to output values close to each  other for both the agent and the players data, while still outputting slightly higher values for the expert-like data.

With no pre-training, $D$ is too easily \textit{tricked}, thus creating random non-informative rewards. With pre-training using a real, low \textit{filled in} value, $D$ is hard to trick. This produces a meaningful reward, but creates a more difficult task for the agent.
However, pre-training with a high \textit{filled in} value reduces the difference between the output of $D$ for the agent and for the expert data. This creates a reward function that is not too \textit{severe} while still giving meaningful feedback to the agent. Using a high \textit{filled in} value can be seen as some form of curriculum learning as it reduces the difference in reward between good and bad behavior, thus creating a smaller slope from bad to good behaviors.

% This allows the agent to perceive meaningful rewards as soon as the training start. This is akin to curriculum learning. Indeed, having a $D$ easily tricked is an easier task than having a $D$ hard to trick, for the generator, thus making the beginning of the training both easier (as $D$ is easily tricked) and more informative (as $D$ still tilts toward player like data).

\subsubsection{Rare D updates and Off-Policy}
RL algorithms require substantial amount of interactions with the environment to converge, especially in a complex setting. On the other hand, training a neural network on a few inputs in a supervised fashion is much faster. This is why $D_\phi$ is trained more rarely than $\pi_\theta$.
Doing so allows the generator to \textit{catch up} with the discriminator. We do not however wait for $\pi_\theta$ to have fully converged, instead a single gradient descent on $D_\phi$ is regularly performed. And because $D_\phi$ is fitted solely on summarized data, it is very cheap to keep the metrics associated with each trajectories generated by the agent in memory, in addition to a classic replay buffer.
This allows the re-use of data generated by $\pi_\theta$ by simply updating its end episode reward signal using the latest version of $D_\phi$.

\section{Experimental Setup}
%\vskip -0.1in
\subsection{Game Environment}
\vskip -0.1in
%The game environment used in this work is the same as in \cite{CARMI}.
The game environment used in this work is developed using Unity's ML-Agents \cite{Unity_Ml_agent}, which allows to control the agent either with the Python programming language (for the RL side) or with a controller (for the human player side).

This environment is a discrete, turn-based, shooter-strategic video game.
It is inspired in its gameplay elements from the "Mario + Rabbids" video game, developed by Ubisoft.
A team of 3 heroes, controlled by an agent (or a human), face a team of a varying number of enemies controlled by behavior trees, for a maximum of 10 turns.
Every hero character can move, shoot and stab. Each of them has a special ability called a \textit{super}. Additionally, each hero has its own stats: health, range of movement, range of attack, damage on attack, so as to differentiate them even more.
Each character, hero and enemy alike, has a maximum of one shot and one stab per turn. Once a character has shot it can not move for the rest of the turn.
We expect this environment to be released to the research community in the coming months.

The environment returns a state vector of size 7414. It is composed of two parts. The first is the flatten representation of the board's ground truth, indicating what object is in each cell (hero, enemy, cover, portal or nothing at all). 
The second part is a vectorial observation, which includes additional information: the number of turns left and the current stats at each hero and enemy.
The number of distinct actions possible at each step is 1258.
Note that not all actions are available for all heroes at all times. For example, some cells might be out of reach for a hero. When an action is unavailable to the agent, it is simply masked, setting to $0$ the probability of its selection.

\subsection{Data Gathering}
\vskip -0.1in
Using Unity's ML-Agents, the data from 25 participants ($N_P$) was gathered. Each played 10 levels. This corresponds roughly to one hour of gameplay. At the end of each level the player would go on to the next level, no matter the outcome.
%The players only knew that this play session was done in order to help automated test, there was no mention of machine learning.
The first level was removed from the available data, as it mainly served as an introduction tutorial level.
%where players where mostly testing the controllers and not engaging fully with the game.
Levels 2 through 9 were kept as a training set ($L_{Train}$) and level 10 as a test level ($L_{Test}$), which means that the training set only contains  $L_{Train} * N_P = 8 * 25 = 200$ sets of aggregated data (i.e. sets of metrics).
The dataset is divided into train and test sets to see if the resulting RL agent is capable of generalizing human-like metrics to new levels. The generalization aspect is often left out of analysis in IRL approaches.

\subsection{Training Procedure}
\vskip -0.1in
The forward RL algorithm used to train the agent is ACER \cite{ACER}. It is a discrete action control algorithm, with an actor-critic architecture. It allows for off-policy updates, fitting well with our approach.
%The off-policy part is coupled with a replay buffer, which is prioritized following \cite{schaul2016prioritized}.
Another reason for choosing ACER is the possibility to run multiple environments in parallel in an asynchronous fashion, all feeding the same buffer and training the same model. 
%This is quite useful for training agents with an environment that is not perfectly stable and could crash.
%"Running multiple environments in parallel" does mean using multiple environments to train the same model, each sharing the same weights and populating the same replay buffer.
The RL neural network architecture is the same as in CARMI, following \cite{Quentin_Catane}: it treats conjointly both vectorial and convolution-based inputs in a way that allows the generated action distribution to also be both vectorial and convolution-based.
Three environments ran in parallel, each training the same model and populating the same replay buffer. Each model trained for the same amount of time. Our setup is a single computer with a 12 core CPU and a NVIDIA GeForce GTX 1070 GPU.

The discriminative model neural network is a series of fully connected layers, taking as input the normalized aggregated data and outputting a probability, following eq. \ref{eq:GAN_R_D_GCL}.

\section{Evaluation And Results}

\begin{table*}
\caption{Jensen–Shannon distance between normalized distribution of players and agents, smaller is better. Comparison showing AILAD ablations of key components.
%Ablations legend available in section \ref{sec:ablation_studies_results}
}{}
\label{tab:ablation_results}
\setlength\tabcolsep{5.5pt}
%\begin{center}
\centerline{
\begin{tabular}{c|c|cccccc|c|}
\cline{2-9}
\multicolumn{1}{l|}{}                                 & \textbf{Metrics}  & \textbf{No $z$} & \textbf{Real Fill In} & \textbf{No Pre-Train} & \textbf{$D_{E}=0$} & \textbf{$D_{E}=1$} & \textbf{No GAN-GCL Corr} & \textbf{AILAD} \\ \hline
\multicolumn{1}{|c|}{\multirow{8}{*}{\rotatebox[origin=c]{90}{\textbf{Train}}}} & Shots             & 0.28                     & 0.45                & 0.32                 & 0.53               & 0.43               & 0.36                 & \textbf{0.28} \\
\multicolumn{1}{|c|}{}                                & Stabs             & \textbf{0.31}            & 0.32                & 0.32                 & 0.58               & 0.36               & 0.36                 & 0.32          \\ \cline{2-9}
\multicolumn{1}{|c|}{}                                & Shots Taken       & 0.31                     & 0.29                & 0.29                 & 0.35               & 0.28               & 0.31                 & \textbf{0.27} \\
\multicolumn{1}{|c|}{}                                & Stabs Taken       & 0.28                     & 0.34                & 0.38                 & 0.35               & 0.34               & 0.34                 & \textbf{0.23} \\ \cline{2-9} 
\multicolumn{1}{|c|}{}                                & Shields           & 0.30                     & 0.29                & 0.26                 & 0.28               & 0.31               & 0.28                 & \textbf{0.24} \\
\multicolumn{1}{|c|}{}                                & Heals             & 0.27                     & 0.28                & 0.28                 & 0.31               & 0.26               & 0.22                 & \textbf{0.23} \\
\multicolumn{1}{|c|}{}                                & Shots   Mpwr   & 0.37                     & 0.39                & 0.37                 & 0.64               & 0.24               & 0.53                 & \textbf{0.19} \\ \cline{2-9} 
\multicolumn{1}{|c|}{}                                & \textbf{All}      & 0.30                     & 0.34                & 0.32                 & 0.43               & 0.32               & 0.34                 & \textbf{0.25} \\ \hline \hline
\multicolumn{1}{|c|}{\multirow{8}{*}{\rotatebox[origin=c]{90}{\textbf{Test}}}}  & Shots             & 0.56                     & 0.69                & 0.67                 & 0.81               & \textbf{0.52}      & 0.62                 & 0.58          \\
\multicolumn{1}{|c|}{}                                & Stabs             & 0.60                     & 0.72                & \textbf{0.54}        & 0.73               & 0.62               & 0.69                 & 0.55          \\ \cline{2-9} 
\multicolumn{1}{|c|}{}                                & Shots Taken       & 0.59                     & 0.64                & \textbf{0.53}        & 0.76               & 0.56               & 0.55                 & 0.57          \\
\multicolumn{1}{|c|}{}                                & Stabs Taken       & 0.52                     & 0.74                & 0.72                 & 0.54               & 0.79               & 0.73                 & \textbf{0.52} \\ \cline{2-9} 
\multicolumn{1}{|c|}{}                                & Shields           & \textbf{0.39}            & 0.50                & 0.49                 & 0.46               & 0.58               & 0.43                 & 0.44          \\
\multicolumn{1}{|c|}{}                                & Heals             & 0.63                     & 0.58                & 0.39                 & 0.52               & 0.58               & 0.51                 & \textbf{0.45} \\
\multicolumn{1}{|c|}{}                                & Shots   Mpwr   & 0.57                     & 0.59                & 0.56                 & 0.66               & 0.61               & 0.61                 & \textbf{0.51} \\ \cline{2-9} 
\multicolumn{1}{|c|}{}                                & \textbf{All}      & 0.55                     & 0.64                & 0.56                 & 0.64               & 0.61               & 0.59                 & \textbf{0.52} \\ \hline
\end{tabular}
}
%\end{center}
%\vskip -0.2in 
\end{table*}

%\vskip -0.1in
%In this section we will compare our approach with different agents. 
Our main baseline is the CARMI \cite{CARMI} agent. The CARMI framework learns a policy $\pi(a|s,z)$ with the following reward $r(\tau) \propto -\Delta(\psi_\mathcal{N}(\tau);z) $. Using normalized metrics in this reward allows to sample $z \sim \mathcal{N}(0,\mathbb{I}^M)$. Thus, sampling $z$ the same way on test levels makes the agent obtain the same metrics as the players.
%who generated the normalization constants on the train levels.
As stated above, no other approaches has been put forward that can apply imitation learning, based on game metrics, on complex environments with a complex relationship between the aggregated data.
Moreover, no other approach takes explicitly into account the potential diversity in the demonstrations.
CARMI takes into account this diversity, and rely solely on aggregated data. Therefore we argue it is the closest approach to ours in terms of constraints and objectives.

We compare AILAD to CARMI. Both models are trained using the same hyperparameters, using the same resources.
Both models are trained on 7 metrics. The number of shots and stabs inflicted on the enemies by the heroes to measure the \textit{aggressiveness} of the play-style. The number of shots and stabs inflicted on the heroes by the enemies to measure the \textit{defensiveness} of the play-style. And finally the number of \textit{shields}, \textit{heals} and \textit{empowers} used by the heroes to measure the \textit{super} usage of the play-style.

The capacity of the agent to match the distribution of the players' metrics  is measured on both the training and test levels. 
For each agent, we report the distance between the data it generates and the players' data, for each of the seven normalized metrics.
The distance used is the Jensen–Shannon divergence ($JSD$), which is a symmetrized and smoothed version of the Kullback–Leibler divergence ($KL$) between two distributions $P$ and $Q$: $JSD(P||Q) =  \frac{KL(P||M) + KL(Q||M)}{2}$ where $M=\frac{P+Q}{2}$. 
Additionally, we perform ablation studies to measure the effect of several key components of our approach.

We report play-style similarity because it is what the agent is judged upon (by the reward function) and it is a good measure of its capability to produce meaningful diversity. Play-styles, as expressed as a collection of game metrics, is the only information available to the algorithm regarding the experts.
Other evaluation metrics, such as reward do no capture this desired diversity. Indeed, this is one of the limitations of existing AIL techniques. Moreover, using synthetic data produced by another agent, to measure the capacity of our agent to retrieve the reward function, is not a good measurement of its capabilities, as demonstrated in \cite{orsini2021matters}.

%In addition to the CARMI agent, we will train an agent whose sole purpose is to win, with a reward of +1 for wins and -1 for losses.

%
%In addition to comparison with the CARMI agent we perform ablation studies on each component detailed in section \ref{sec:stbaility_inprov}. 

%\subsection{Baselines}
%CARMI  ==> une approche goal conditioned qui train sur les meme metrics arrivent à atteindre quoi avec une reward linéaire faite à la main ?
%win only ==> avoir une idée de à quel point la stochasticité de l'env fait le taff à notre place: un training sans but de mapper une distribution est divers à quel point ?

\begin{table*}[h]
\caption{Average in-game metrics values for players and for CARMI and AILAD models. The upper part of the table represents the metrics used during the training of the agents. The lower part are metrics which were not used during the training of the agents.}{}
\label{tab:absolute_metrics}
%\vskip 0.15in
\centerline{
\begin{tabular}{c|ccc|ccc|}
\cline{2-7}
                                                                                                   & \multicolumn{3}{c|}{\textbf{Train}}                & \multicolumn{3}{c|}{\textbf{Test}}                 \\ \hline
\multicolumn{1}{|c|}{\textbf{Metrics}}                                                             & \textbf{Players} & \textbf{CARMI} & \textbf{AILAD} & \textbf{Players} & \textbf{CARMI} & \textbf{AILAD} \\ \hline
\multicolumn{1}{|c|}{Shots}                                                                        & 1.9              & 1.2            & 1.7            & 2.1              & 1.5            & 1.8            \\
\multicolumn{1}{|c|}{Stabs}                                                                        & 1.1              & 0.8            & 1.0            & 1.3              & 1.1            & 1.4            \\
\multicolumn{1}{|c|}{Shots Taken}                                                                  & 0.3              & 0.3            & 0.3            & 0.4              & 0.3            & 0.3            \\
\multicolumn{1}{|c|}{Stabs Taken}                                                                  & 0.2              & 0.3            & 0.2            & 0.2              & 0.2            & 0.2            \\
\multicolumn{1}{|c|}{Shields}                                                                      & 0.1              & 0.2            & 0.1            & 0.2              & 0.2            & 0.1            \\
\multicolumn{1}{|c|}{Heals}                                                                        & 0.2              & 0.2            & 0.2            & 0.3              & 0.3            & 0.2            \\
\multicolumn{1}{|c|}{Shots Empowered}                                                              & 0.5              & 0.4            & 0.5            & 0.6              & 0.6            & 0.4            \\ \hline
\multicolumn{1}{|c|}{Average Distance Between   Heroes} & 4.7              & 7.9            & 4.6            & 5.0              & 6.7            & 3.7            \\
\multicolumn{1}{|c|}{Average Distance To Enemies}        & 8.3              & 9.8            & 8.5            & 8.4              & 10.4           & 8.5            \\
\multicolumn{1}{|c|}{Heroes \% Lost HP}                                                            & 31               & 72             & 55             & 39               & 73             & 70             \\
\multicolumn{1}{|c|}{Enemies \% Lost HP}                                                           & 98               & 75             & 91             & 98               & 75             & 94             \\
\multicolumn{1}{|c|}{\% Win}                                                                       & 89               & 20             & 66             & 91               & 7              & 56             \\
\multicolumn{1}{|c|}{\% Lost}                                                                      & 0                & 22             & 10             & 0                & 30             & 9              \\
\multicolumn{1}{|c|}{\% Draw}                                                                      & 11               & 58             & 24             & 9                & 63             & 35             \\
\multicolumn{1}{|c|}{Nb of Turns}                                                                  & 6.5              & 9.1            & 7.8            & 7.5              & 9.2            & 9.0            \\ \hline
\end{tabular}
}
\end{table*}

\subsection{Expert Distribution Mapping}
%\vskip -0.1in
After both models, CARMI and AILAD, have been trained, they are used to play both train and test levels multiple times, each time with a different randomly selected $z$.
We report the JSD between each model and the experts distribution. The results are reported in Table \ref{tab:CARMI_comparison}. 
Each model has been trained using the same expert data.
%AILAD is more expensive to run, due to the discriminative neural network. This additional network takes only the aggregated data as input, thus has only a few parameters to train and does not require a lot of additional compute power.
In the interest of fairness, both models have been trained for the same amount of time. 

%\vskip -0.3in
\begin{table}
\caption{Jensen–Shannon distance between normalized distribution of players and agents, smaller is better.
Comparison between CARMI and AILAD agents.}{}
\label{tab:CARMI_comparison}
%%\vskip -0.5in
\begin{center}
\setlength\tabcolsep{5.5pt}
%\begin{small}
%\begin{sc}
\begin{tabular}{c|cc|cc|}
\cline{2-5}
\multicolumn{1}{l|}{}                 & \multicolumn{2}{c|}{\textbf{Train}}       & \multicolumn{2}{c|}{\textbf{Test}}        \\ \cline{2-5} 
                                      & CARMI                     & AILAD          & CARMI                     & AILAD          \\ \hline
\multicolumn{1}{|c|}{Shots}           & \multicolumn{1}{c|}{0.52} & \textbf{0.28} & \multicolumn{1}{c|}{0.69} & \textbf{0.58} \\
\multicolumn{1}{|c|}{Stabs}           & \multicolumn{1}{c|}{0.32} & \textbf{0.32} & \multicolumn{1}{c|}{0.59} & \textbf{0.55} \\ \hline
\multicolumn{1}{|c|}{Shots Taken}     & \multicolumn{1}{c|}{0.29} & \textbf{0.27} & \multicolumn{1}{c|}{0.62} & \textbf{0.57} \\
\multicolumn{1}{|c|}{Stabs Taken}     & \multicolumn{1}{c|}{0.31} & \textbf{0.23} & \multicolumn{1}{c|}{0.57} & \textbf{0.52} \\ \hline
\multicolumn{1}{|c|}{Shields}         & \multicolumn{1}{c|}{0.29} & \textbf{0.24} & \multicolumn{1}{c|}{0.48} & \textbf{0.44} \\
\multicolumn{1}{|c|}{Heals}           & \multicolumn{1}{c|}{0.28} & \textbf{0.23} & \multicolumn{1}{c|}{0.52} & \textbf{0.45} \\
\multicolumn{1}{|c|}{Shots Empowered}   & \multicolumn{1}{c|}{0.3}  & \textbf{0.19} & \multicolumn{1}{c|}{0.63} & \textbf{0.51} \\ \hline
\multicolumn{1}{|c|}{\textbf{All}}    & \multicolumn{1}{c|}{0.33} & \textbf{0.25} & \multicolumn{1}{c|}{0.59} & \textbf{0.52} \\ \hline
\end{tabular}
%\end{sc}
%\end{small}
\end{center}
%\vskip -0.3in
\end{table}

AILAD clearly out performs CARMI, on each metric and on both train and test levels. One result worth noting is the fact that AILAD produced roughly similar results on each metrics, while CARMI is more inconsistent. This is due to the non-linear nature of AILAD's reward function. Indeed, in this adversarial framework, if the agent were to perform particularly poorly on one specific metric, then the discriminative model would only need to focus on this particular metric to tell the difference between the data from the expert and from the agent. On the other hand, with CARMI, performing badly on one metric can only penalize the agent up to a point.

\subsection{Ablation Studies}\label{sec:ablation_studies_results}
\vskip -0.1in

In this section the AILAD model is compared to 6 ablated version of itself.
The benefice of each of the component of AILAD is measured: the key ones as well as the ones used mainly for stability in the training. The JSD between each of the models and the players is reported on both train and test levels in Table \ref{tab:ablation_results}. Overall, the \textit{complete} AILAD agent outperforms all. It does so by quite a margin. This proves the importance of each of these component.

The first model is trained without a latent space $z$. Without this additional input the agent has no obvious way to produce diversity. It shows that adding a latent space does in fact improve the performance of the model, in terms of matching the desired distribution of experts.
Moreover, this corresponds to a naive implementation of the GAN-GCL algorithm, without any of the novelty introduced in our work.

During the pre-training of the discriminative model, detailed in section \ref{sec:stbaility_inprov}, some pre-computed value must be filled in, in eq. \ref{eq:GAN_R_D_GCL}. In AILAD, this value is fixed to 0.3 during the pre-training. Instead in the second model, the real density of the synthetic data used to generate the fake examples is used.
As described, it produces a more \textit{severe} reward function, making the learning of the agent harder in the beginning.
This shows that it is better to start the training with a more lenient reward function, not penalizing bad behaviors too harshly.

The third model however makes no pre-training of $D$ whatsoever. Doing so means that during the early stages of the agent training, the reward function is not well adjusted.
The feedback given to the agent through that reward function is then mostly noise. Re-adjusting its behavior afterwards is challenging and the agent might not recover from such a poor start.

The next two models used different values for the hyperparameters $D_{E}$ which is used to know with which frequency to update the discriminative network. Setting $D_{E}$ to $0$ corresponds to never updating it. It is still being pre-trained though.
This demonstrates the advantages of the adversarial framework. This model still has the non linearity of the reward function, however it performs even worse than CARMI. This shows that a non linear reward function is better only if it is trained conjointly with the agent.
On the other hand, $D_{E}=1$ means that the discriminative model is trained with the same frequency as the generator. As stated in section \ref{sec:stbaility_inprov}, having an ever changing reward function is no easy task for RL agents. This is reflected in the results of this model.

And finally, the last model is one where no particular structure is imposed on the discriminative model outputs, as in GAIL: it is simply a sigmoid output. 
%Removing this correction to $D$'s outputs transform this problem back into a classic GAN problem, where no point of convergence exists. 
Here, it translates into having a reward function that does not tend toward a single optimum but fluctuates during training, and as with the previous model, a fluctuating reward function impacts negatively the performances of any RL agent.

\subsection{In-Game Metrics}\label{sec:Appendix_absolute_mretrics_reporting}
\vskip -0.05in
In this section, we report the average of each metrics, and other not used during the training, for the players, our model AILAD and our baseline CARMI. These results are displayed in Table \ref{tab:absolute_metrics}. This is what models like CARMI or AILAD would be used to produce. In this context of anticipating players behaviors on new levels, AILAD would produce more precise results. These results show again that AILAD is closer to the players than CARMI is.

Moreover, we report metrics that were not used during the training (the second part of the the table).  Even on these metrics AILAD is better than CARMI. This is due to the fact that AILAD must take into account the relationship between each metrics, rather than treating them independently. So the more constrained AILAD is on the training metrics, the less \textit{liberty} to deviate from players it has on metrics that were not part of the reward function. Notably the "Average Distance Between Heroes", which measures how grouped or how scattered is the blue team, and the "Average Distance to Enemies" of AILAD, which measures how close the blue team gets to the red team, is consistent with the players'. 

Another thing worth noting in this table are the win rates. While performing better than CARMI, AILAD is not at players' level yet. However, unlike CARMI, the "Enemies \% Lost HP" of AILAD, which measure how closely the agent was to defeating the enemies, is almost as high as the players'. This could mean that the solution to player-like win rates would be to add more metrics relative to defensive aspect of the game.

The results presented here, serve to illustrate how AILAD and CARMI might be used in the video game industry for example, and how AILAD produce more meaningful feedback than CARMI. Table \ref{tab:absolute_metrics} is an example of the type of feedback these models would be used to produce in a video game production team. Video game production is a context that fits well with the constraints used in these approaches, the data would correspond to that which is obtained through play-test. Indeed, during the production of video games, and even more so during the early stages, the play-tests put in place result in low quantity data, usually in the form of aggregated data, with ideally a wide range of play-styles.

\section{Conclusion}

We presented AILAD, an agent capable of generating diverse behaviors producing a distribution of expert-like aggregated data, through an adversarial framework. 
This approach does not suffer from the three main IRL constraints: having a RL solver in the inner loop, access to full trajectories and homogeneity in the expert demonstrations.
Indeed this approach learns conjointly the reward function and the optimal policy, making it scalable to complex environments. It does so, not requiring full paths which are expensive to gather, but aggregated data, making it more applicable to real-world problems.
Finally, it generates diversity so as to produce behaviors reflecting a pool of experts.
Moreover, we presented several ways to improve the performances of the model, and measured their impact on performances, through a extensive ablation study.

However, few guarantees can be made on overall agent behavior matching the experts. Nonetheless, in certain situations, the metrics generated are more important than the behavior generating them, for instance when performing automated tests on games, where the goal is to provide aggregated feedback to game designers about players' performance within the game.

%Moreover, we presented in this work many improvements which allows such unstable models, as GANs and IRL, to converge nicely to a desired output.
Our work combines advantages of the AIL framework (non linear reward function) and of the CARMI approach (relying on aggregated data with a focus on diversity), in a single powerful solution and adds convergence tricks to create the first working imitation learning model from aggregated data using a dynamically adaptive reward function.

%Because this approach recovers the latent reward function, this could be useful to train $\pi$ in contexts where no expert data has been gathered yet for example. This could improve its generalization capabilities.
%Going forward, it would be interesting to combine this approach with recent advances in agents which generate a distribution of trajectories, proportional to their respective rewards, such as G-Flow Network \cite{FlowNetwork}.

\bibliography{ecai}

\end{document}